\documentclass{article}

\usepackage[preprint]{neurips_2025}

\usepackage{multirow}
\usepackage{graphicx}
\usepackage[utf8]{inputenc} 
\usepackage[T1]{fontenc}    
\usepackage{hyperref}       
\usepackage{url}            
\usepackage{booktabs}       
\usepackage{amsfonts}       
\usepackage{nicefrac}       
\usepackage{microtype}      
\usepackage[table]{xcolor}         

\usepackage{enumitem}

\title{R2SM: Referring and Reasoning for Selective Masks}

\author{%
  Yu-Lin Shih$^{1}$\thanks{These authors contributed equally to this work.}\enspace, Wei-En Tai$^{1}$\footnotemark[1]\enspace,
  Cheng Sun$^2$, Yu-Chiang Frank Wang$^{2,3}$, Hwann-Tzong Chen$^1$ \\
  $^1$National Tsing Hua University\\
  $^2$NVIDIA, $^3$National Taiwan University\\
}

\begin{document}

\maketitle

\begin{abstract}

We introduce a new task, \textit{Referring and Reasoning for Selective Masks (R2SM)}, which extends text-guided segmentation by incorporating \textit{mask-type selection} driven by user intent. This task challenges vision-language models to determine whether to generate a \textit{modal} (visible) or \textit{amodal} (complete) segmentation mask based solely on natural language prompts. To support the R2SM task, we present the \textit{R2SM dataset}, constructed by augmenting annotations of COCOA-cls, D2SA, and MUVA. The R2SM dataset consists of both modal and amodal text queries, each paired with the corresponding ground-truth mask,
enabling model finetuning and evaluation for the ability to segment images as per user intent.
Specifically, the task requires the model to interpret whether a given prompt refers to only the visible part of an object or to its complete shape, including occluded regions, and then produce the appropriate segmentation. For example, if a prompt explicitly requests the whole shape of a partially hidden object, the model is expected to output an amodal mask that completes the occluded parts. In contrast, prompts without explicit mention of hidden regions should generate standard modal masks. The R2SM benchmark provides a challenging and insightful testbed for advancing research in multimodal reasoning and intent-aware segmentation.

\end{abstract}
\begin{figure}[ht]
  \centering
  \includegraphics[width=\linewidth]{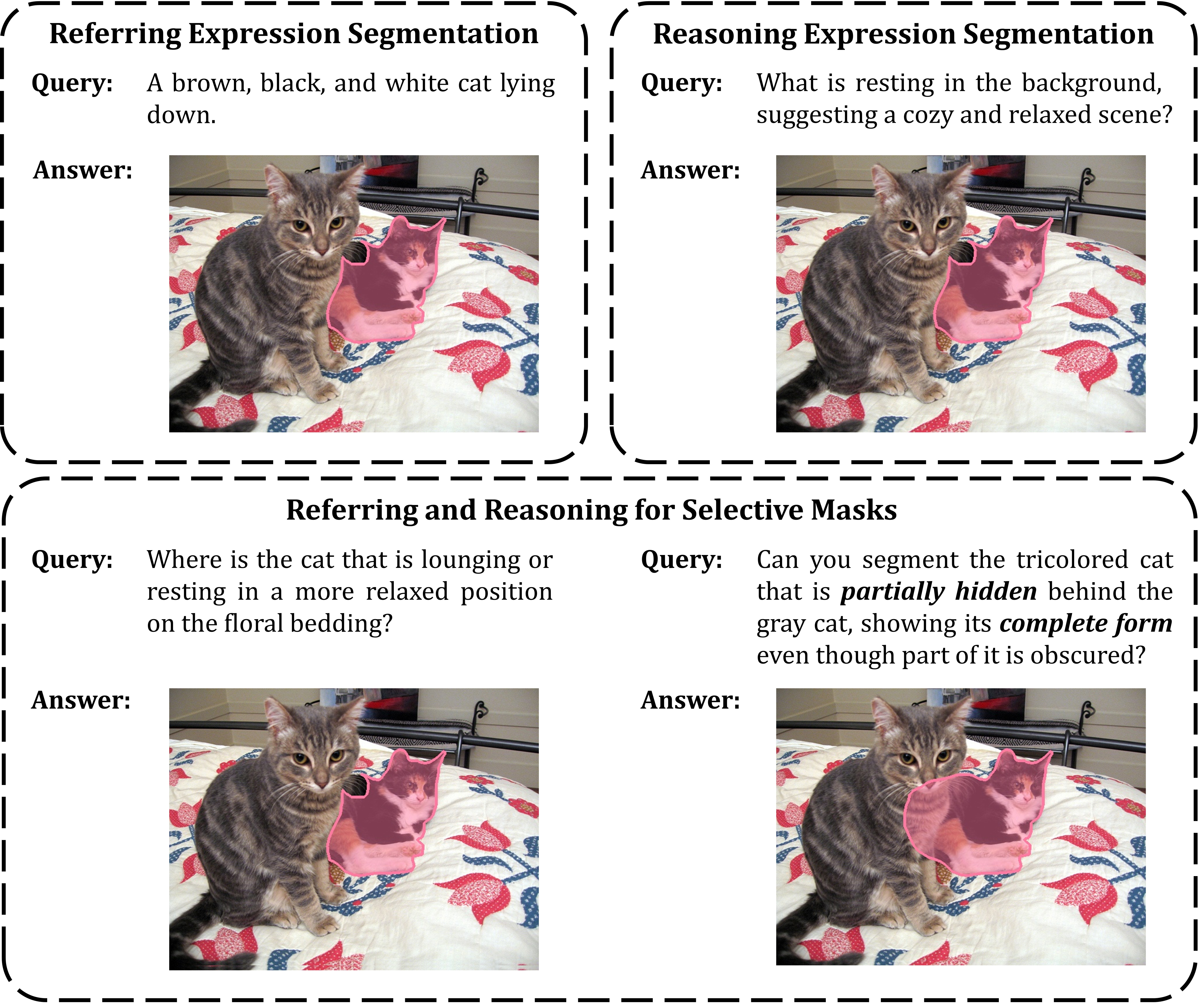}
  \caption{Comparison of task formulation among Referring Expression Segmentation, Reasoning Expression Segmentation, and Referring and Reasoning for Selective Masks (R2SM).}
  \label{fig:vis_task}
\end{figure}
\section{Introduction}
\subsection{Reasoning Segmentation and Occlusion Understanding}

Recent advances in vision-language models (VLMs) have led to a growing number of systems that combine large language models (LLMs) with visual perception capabilities. One of the key downstream tasks is 
image segmentation given natural language description.
This may include two types of segmentation:
{\it i)} referring segmentation, which uses short and precise prompts, and {\it ii)} reasoning segmentation,
which involves more complex or indirect descriptions that require contextual understanding.

Most existing models have limited capabilities in handling occlusion. They often fail to infer the complete shape of a partially occluded object, even when the prompt explicitly requests segmentation of the entire object. While some recently developed VLMs like AURA~\cite{abs-2503-10225}
are specialized to generate amodal segmentations, they lack the ability to decide whether to produce a modal or amodal mask based on user input. The idea of allowing users to select the type of mask remains underexplored and lacks dedicated benchmarks, which motivates our work.

This paper introduces a new task called \textit{Referring and Reasoning for Selective Masks (R2SM)}, which extends the text-guided segmentation paradigm by adding an extra layer of complexity. The model must not only interpret the meaning of the input text but also \textit{select} whether to segment the visible (modal) or the complete (amodal) region of the object, purely based on the user's natural language prompt (see Figure~\ref{fig:vis_task}). This capability is relevant to applications such as assistive robotics, autonomous driving, augmented reality, and medical imaging, where intent-aware segmentation and occlusion reasoning are essential.


\subsection{Limitations of Existing Models and Amodal Datasets}

The amodal segmentation task receives extensive attention in the vision community. Several datasets, such as COCOA-cls~\cite{FollmannKHKB19}, D2SA~\cite{FollmannKHKB19}, MUVA~\cite{LiYTBZJ023}, and Amodal-LVIS~\cite{abs-2503-06261}, provide pixel-level amodal annotations for every instance in an image. These datasets primarily support full-scene amodal segmentation or instance-level amodal completion, serving as essential resources for training models to recover occluded object shapes. However, these datasets provide only class labels as textual annotations, lacking more complex natural language prompts required by the emerging class of vision-language models.

On the modeling side, VLMs such as LISA~\cite{LaiTCLY0J24} and PSALM~\cite{ZhangMZB24} have demonstrated strong performance in general image-text understanding tasks. Despite this, their ability to reason about occlusion remains limited. Without paired text and amodal annotations, these models do not have the necessary supervision to handle prompts that involve occluded content. As a result, they are incapable of generating amodal masks and therefore cannot determine when to produce modal or amodal segmentations based on linguistic cues. 

To address these limitations, we introduce the \textit{R2SM dataset}~\footnote{The dataset is available at \url{https://huggingface.co/datasets/vllabnthu/R2SM}}, a new benchmark designed specifically for this task.

\subsection{Our Contributions}
Our key contributions can be summarized as follows:
\begin{itemize}[leftmargin=8mm,label=$\circ$]
    \item We introduce \textit{Referring and Reasoning for Selective Masks}, a new task that requires models to infer user intent and choose between modal and amodal segmentation based on natural language prompts.
    \item We present the \textit{R2SM dataset}, the first reasoning segmentation dataset that combines both modal and amodal text prompts with corresponding mask annotations. This dataset enables fine-grained evaluation of occlusion reasoning and intent understanding.
    \item We conduct comprehensive experiments on several representative VLMs, analyzing their ability to handle both modal and amodal prompts. Our analysis identifies their strengths and weaknesses and discusses potential improvements for future work in reasoning-based segmentation.
\end{itemize}

\section{Related Work}
\subsection{Pixel-level Vision-language Models}
Vision-language models have increasingly been applied to fine-grained pixel-level tasks such as segmentation, grounding, and referring comprehension. Early works like RefTR~\cite{LiS21}, SeqTR~\cite{ZhuZSLPLCCSJ22}, and LAVT~\cite{0002WT0ZT22} focus on Referring Expression Segmentation (RES) in a single-object setting. GRES~\cite{0072DJ23} introduces Generalized RES, which supports expressions that refer to multiple targets or even no targets at all.

Beyond RES, recent models integrate multimodal LLMs to tackle more general and open-ended pixel-level reasoning. LISA~\cite{LaiTCLY0J24} proposes an embedding-as-mask mechanism for complex language queries, while its successor, LISA++~\cite{abs-2312-17240}, expands segmentation capabilities to multi-object scenarios through instance-aware supervision. Other models, including PixelLM~\cite{RenHW0FFJ24}, GLaMM~\cite{Rasheed0MS0CAX024}, SAM4MLLM~\cite{ChenLSWC24}, OMG-LLaVA~\cite{ZhangL0YWJLY24}, and PSALM~\cite{ZhangMZB24}, further consolidate diverse vision-language tasks into unified frameworks. Additionally, AURA~\cite{abs-2503-10225} is the first to introduce amodal reasoning segmentation, enabling the model to infer occluded regions.

\subsection{Text-guided Segmentation Datasets}
Text-guided segmentation evolves from referring expression tasks to more generalized and open-vocabulary settings. Early datasets like RefCOCO~\cite{YuPYBB16}, RefCOCO+~\cite{YuPYBB16}, and RefCOCOg~\cite{MaoHTCY016} focus on segmenting single objects based on concise phrases. Later, datasets such as PhraseCut~\cite{WuLCBM20} and gRefCOCO~\cite{0072DJ23} extend this to multi-object and zero-object conditions. For broader linguistic grounding, Flickr30k Entities~\cite{PlummerWCCHL15} offers phrase-level supervision through region annotations, while PACO~\cite{RamanathanKPWZG23} provides detailed part and attribute labels. More recently, datasets like ReasonSeg~\cite{LaiTCLY0J24} enable pixel-level understanding under complex linguistic prompts, while ReasonSeg-Inst~\cite{abs-2312-17240} and AmodalReasonSeg~\cite{abs-2503-10225} introduce multi-object reasoning and amodal segmentation. Concurrently, GranD~\cite{Rasheed0MS0CAX024} supports dense region-text alignment, and MUSE~\cite{RenHW0FFJ24} benchmarks multi-target segmentation with diverse visual reasoning capabilities.
Existing datasets are dedicated to either modal or amodal masks, whereas only our dataset considers the ability to switch between modes.

\subsection{Amodal Datasets}
Amodal segmentation research relies heavily on high-quality annotated datasets. As the first dataset specifically designed for this task, COCOA~\cite{ZhuTMD17} provides human-annotated semantic-level masks based on COCO~\cite{LinMBHPRDZ14} images. COCOA-cls and D2SA~\cite{FollmannKHKB19} extend this foundation with instance-level annotations and class labels; D2SA is based on synthetic images from D2S~\cite{FollmannBHKU18} dataset, while COCOA-cls remains manually labeled. MUVA~\cite{LiYTBZJ023} introduces a large-scale synthetic multi-view dataset that focuses on shopping scenes.

In addition, several datasets contribute complementary perspectives to the field: DYCE~\cite{EhsaniMF18} contains synthetic indoor scenes, KINS~\cite{QiJ0SJ19} features manually labeled traffic environments, and WALT~\cite{ReddyTN22} uses time-lapse imagery to infer amodal annotations. MP3D-Amodal~\cite{ZhanZXZ24} is derived from real-world indoor scans in the Matterport3D~\cite{ChangDFHNSSZZ17} dataset. Pix2gestalt~\cite{OzgurogluLS0DTV24} provides synthetic occlusion data for generative amodal completion. Collectively, these datasets enrich the field with diverse annotation methodologies and domain coverage.

\section{Task Formulation}
\subsection{Referring Segmentation}
Referring segmentation is a vision-language task that aims to segment the object referred to by a natural language expression in an image. This task requires the model to understand the linguistic content of the input text and identify the target object accordingly. As illustrated in Figure~\ref{fig:vis_task}, given the prompt \textit{``A brown, black, and white cat lying down''}, the model should generate a segmentation mask that precisely outlines the mentioned cat. Referring segmentation sees extensive development across various datasets and models, with a focus on improving grounding accuracy and segmentation quality. However, these models typically assume that the segmentation target corresponds only to the visible portion of the object.

\subsection{Reasoning Segmentation}
Reasoning segmentation extends the referring task by requiring the model to perform additional inference or commonsense reasoning based on the user prompt. Instead of directly describing the object, the prompt often implies the target through functional or contextual cues. As shown in Figure~\ref{fig:vis_task}, given the prompt \textit{``What is resting in the background, suggesting a cozy and relaxed scene?''}, the model should identify and segment the cat as the correct object, even though the word \textit{``cat''} is not explicitly stated. This category of segmentation demands a deeper understanding of the scene, progressing from simple spatial grounding to more complex semantic and temporal inference. 

Recent models like AURA~\cite{abs-2503-10225} further advance reasoning segmentation by enabling amodal mask prediction, allowing the model to segment objects beyond their visible regions. Nevertheless, these models still lack the capacity to \textit{decide} whether to generate a modal or amodal mask based on user intent. They assume a fixed mask type rather than dynamically adapting to the prompt. In contrast, our approach offers the flexibility to select the mask type based on the user's intent through prompts.

\subsection{Referring and Reasoning for Selective Masks}
We propose \textit{Referring and Reasoning for Selective Masks (R2SM)}, a new task that combines referring and reasoning segmentation while introducing a novel challenge: \textit{selecting} whether to produce a modal or amodal mask based on user intent. The model must interpret the language prompt to decide whether to segment only the visible part of the object or its full shape, including occluded regions. As demonstrated in the bottom row of Figure~\ref{fig:vis_task}, when the user prompt includes phrases such as \textit{``the complete form of the cat''} or \textit{``even the part hidden behind the gray cat''}, the model is expected to infer an amodal mask. Conversely, if the prompt does not refer to occlusion or hidden parts, the model defaults to predicting a modal mask. The task evaluates the ability of a model to align visual reasoning with a nuanced language understanding in scenarios involving occlusion, supporting a more flexible and intent-aware approach to segmentation.

\begin{figure}[ht]
  \centering
  \includegraphics[width=\linewidth]{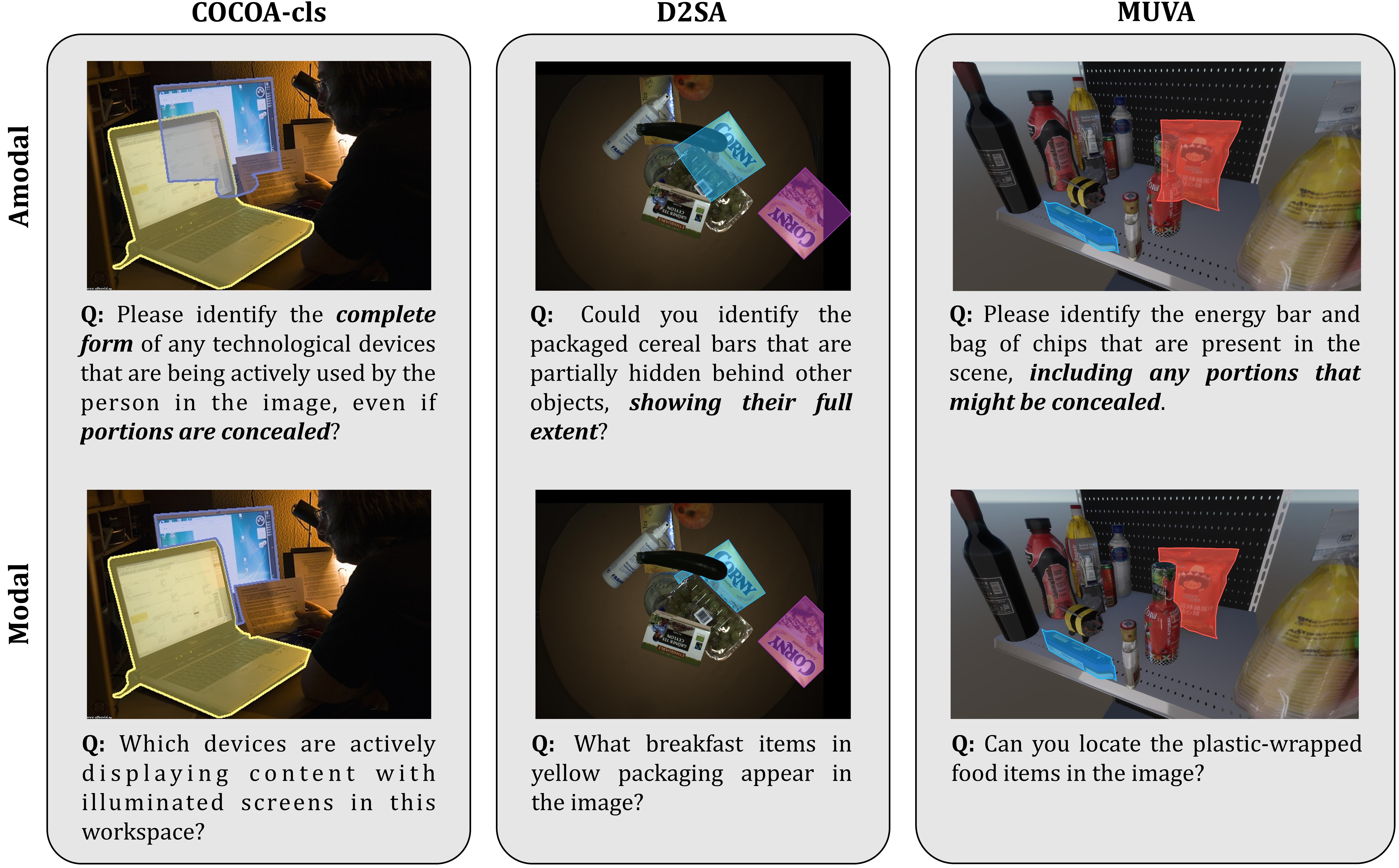}
  \caption{Examples from the R2SM dataset with modal and amodal queries across COCOA-cls, D2SA, and MUVA splits.}
  \label{fig:vis_dataset}
\end{figure}
\begin{figure}[ht]
  \centering
  \includegraphics[width=\linewidth]{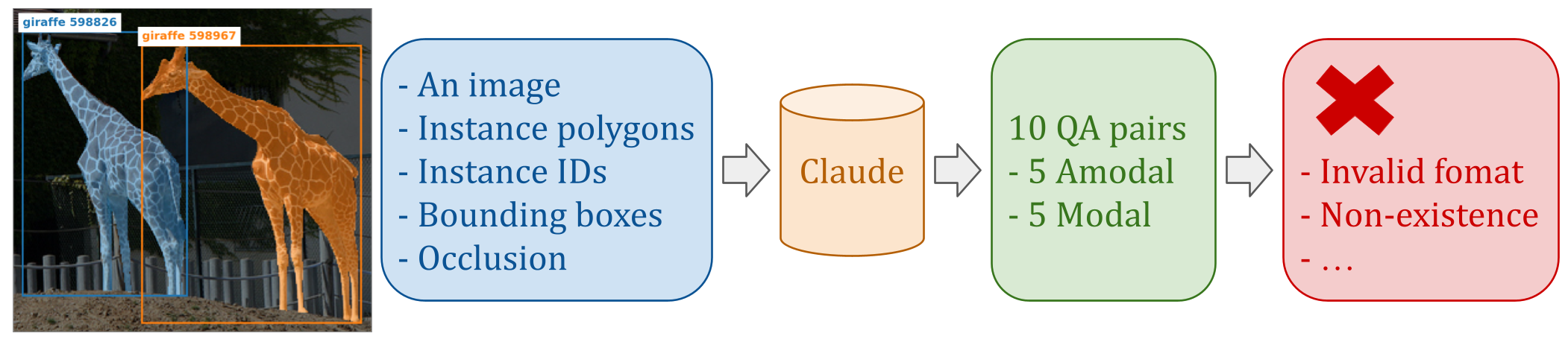}
  \caption{Text query generation pipeline of the R2SM dataset.}
  \label{fig:vis_pipeline}
\end{figure}
\section{R2SM Dataset}
We introduce the \textit{R2SM dataset}, the first dataset designed to support referring and reasoning for selective masks. The R2SM dataset provides paired modal and amodal language prompts with corresponding segmentation masks, allowing models to infer the appropriate mask type based on user intent. Each image is annotated with up to 10 diverse text queries that explicitly or implicitly suggest the target mask type. These queries are grounded with segmentation masks corresponding to either the modal or amodal object regions. The R2SM dataset is built by augmenting existing amodal segmentation datasets, including COCOA-cls, D2SA, and MUVA, as shown in Figure~\ref{fig:vis_dataset}.

\subsection{Source Datasets}
COCOA-cls consists of real-world images with rich contextual scenes, providing amodal mask annotations for objects across a wide range of indoor and outdoor settings. We include all 2,276 training and 1,223 validation images from COCOA-cls as the R2SM COCOA-cls split.

D2SA consists of semi-synthetic images that feature a variety of product categories such as tea, apples, and other textured packaging. Objects are arranged to create realistic occlusions on flat surfaces. We include 2,000 training and 3,600 validation images as the R2SM D2SA split.

MUVA follows a similar design philosophy to D2SA, focusing on household goods. However, MUVA introduces more diversity in photo angles and layouts, simulating more complex occlusion scenarios. From MUVA, we select 6,000 images, divided into 3,000 for training and 3,000 for validation, forming the R2SM MUVA split.

\subsection{Text Query Generation}
Figure~\ref{fig:vis_pipeline} shows the automatic text query generation pipeline. We utilize the Claude API~\footnote{\url{https://www.anthropic.com/api}}, employing a methodology similar to the dataset generation process in \cite{abs-2312-17240}. Specifically, for each image from the source datasets, we provide Claude with comprehensive annotations, including instance mask polygons, instance IDs, bounding box coordinates, and category names.

The language model is prompted to generate diverse questions and statements that extend beyond mere visual appearance, explicitly considering the functional properties and natural context of objects, as well as capturing relational dynamics between multiple instances within each image. Each generated textual query may reference one or more object instances and can include multiple categories within a single query.

In addition, we provide Claude with occlusion information for instances. If an instance is marked as \textit{"is\_occluded"}, Claude generates an additional amodal variant of the text query, explicitly mentioning that the instance is occluded and requesting the complete or inferred outline of the hidden parts. These amodal text queries are then paired with their corresponding amodal mask annotations.

During the query generation process, we enhance the accuracy of Claude’s output by providing illustrative examples. Specifically, we input example instance annotations along with sample textual queries, which significantly reduce errors and improve the overall quality of the generated prompts. Detailed examples are provided in the appendix.

After the generation of text queries, we perform a cleaning step to remove low-quality samples. These include queries with invalid formats, references to non-existent instances, or cases where annotations do not contain the specified instance IDs.

\subsection{Dataset Statistics}
Our COCOA-cls split contains a total of 34,904 text queries, with 22,697 samples in the training set and 12,207 in the validation set. The D2SA split includes 53,783 queries, with 17,784 for training and 35,999 for validation. The MUVA split provides 59,996 queries, evenly divided into 29,998 training and 29,998 validation samples. In each split, approximately half of the text prompts are amodal, while the other half are modal, enabling balanced evaluation for both segmentation types. 
The detailed distribution of the R2SM dataset is summarized in Table~\ref{tab:dataset_tab}.

\begin{table}[h]
\caption{R2SM dataset statistics. Number of images, instances, and text queries for each split. \textit{Amodal} (\checkmark) denotes amodal-only queries; ($\times$) denotes modal-only queries.}
\label{tab:dataset_tab}
\centering
\begin{tabular}{lccccccc}
\toprule
\multirow{2}{*}{\textbf{Split}} &\multicolumn{2}{c}{\textbf{\hspace{6pt}\# Images}} & \multicolumn{2}{c}{\textbf{\hspace{6pt}\# Instances}}& \multirow{2}{*}{\textbf{\hspace{6pt}Amodal}}  & \multicolumn{2}{c}{\textbf{\hspace{6pt}\# Text queries}}  \\
  &\hspace{6pt}Train & Val & \hspace{6pt}Train & Val & \hspace{6pt} & \hspace{6pt}Train & Val\\

\toprule
\multirow{2}{*}{COCOA-cls}  & \hspace{6pt}\multirow{2}{*}{2,276} & \multirow{2}{*}{1,223} &  \hspace{6pt}\multirow{2}{*}{6,763}  & \multirow{2}{*}{3,799} & \hspace{6pt}\checkmark & \hspace{6pt}11,338 & 6,095 \\
                            & \hspace{6pt}                       &                        &  \hspace{6pt}                        &                        & \hspace{6pt}$\times$   & \hspace{6pt}11,359 & 6,112 \\
\midrule
\multirow{2}{*}{D2SA}       & \hspace{6pt}\multirow{2}{*}{2,000} & \multirow{2}{*}{3,600} &  \hspace{6pt}\multirow{2}{*}{13,066} & \multirow{2}{*}{15,654}& \hspace{6pt}\checkmark & \hspace{6pt}9,987 & 18,000 \\
                            & \hspace{6pt}                       &                        &  \hspace{6pt}                        &                        & \hspace{6pt}$\times$   & \hspace{6pt}7,797 & 17,999 \\
\midrule
\multirow{2}{*}{MUVA}       & \hspace{6pt}\multirow{2}{*}{3,000} & \multirow{2}{*}{3,000} &  \hspace{6pt}\multirow{2}{*}{22,246} & \multirow{2}{*}{22,674}& \hspace{6pt}\checkmark & \hspace{6pt}14,999 & 15,000 \\
                            & \hspace{6pt}                       &                        &  \hspace{6pt}                        &                        & \hspace{6pt}$\times$   & \hspace{6pt}14,999 & 14,998 \\
\toprule
\multirow{2}{*}{Total}      & \hspace{6pt}\multirow{2}{*}{7,276} & \multirow{2}{*}{7,823} &  \hspace{6pt}\multirow{2}{*}{42,075} & \multirow{2}{*}{42,127}& \hspace{6pt}\checkmark & \hspace{6pt}36,324 & 39,095 \\
                            & \hspace{6pt}                       &                        &  \hspace{6pt}                        &                        & \hspace{6pt}$\times$   & \hspace{6pt}34,155 & 39,109 \\
\bottomrule
\end{tabular}
\end{table}
\begin{figure}[ht]
  \centering
  \includegraphics[width=\linewidth]{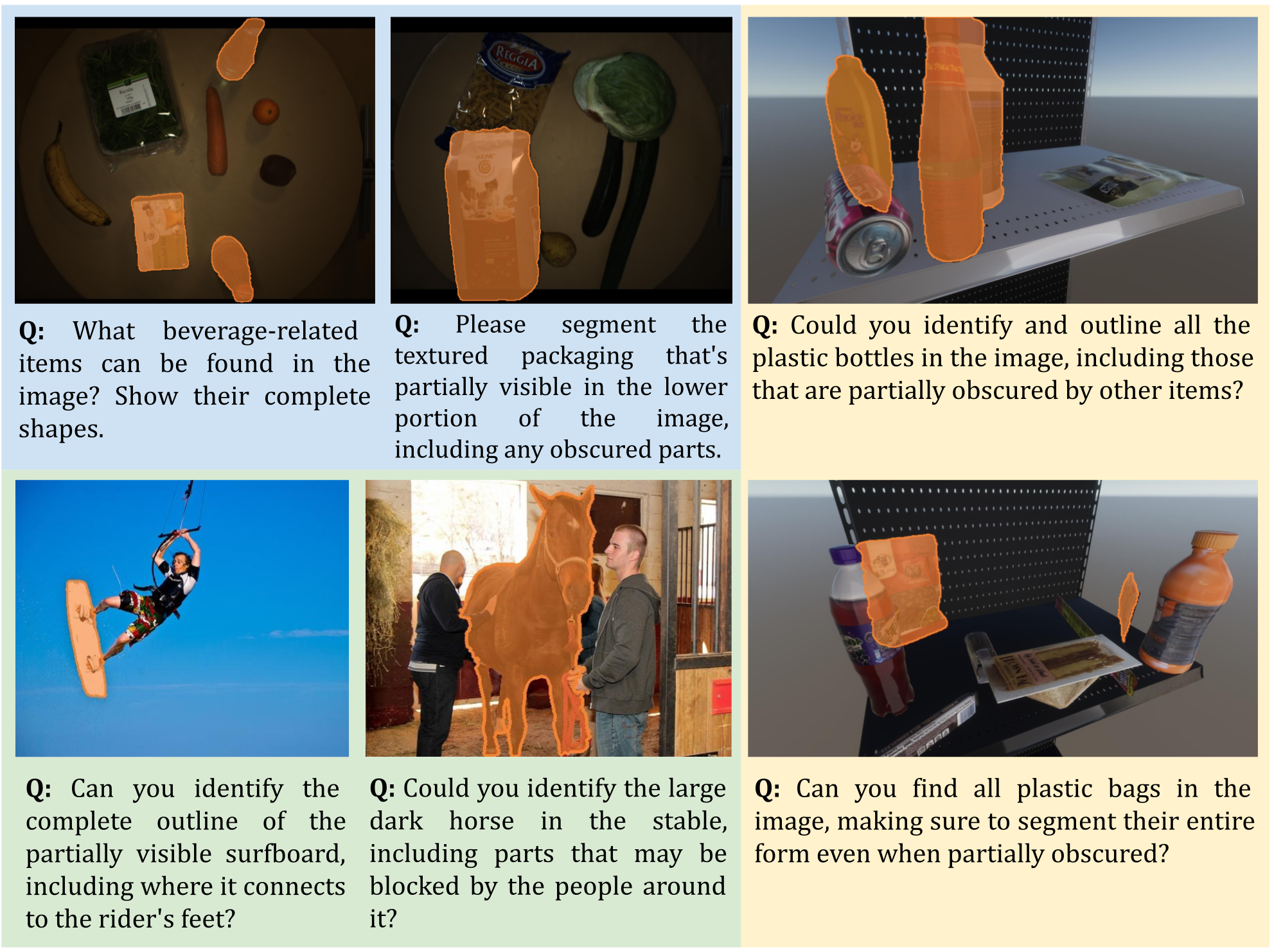}
  \caption{Visualizations on R2SM dataset splits using the best fine-tuned models: GLaMM for COCOA-cls (green), ReLA for D2SA (blue) and MUVA (yellow).}
  \label{fig:vis_models}
\end{figure}
\section{Model Evaluation}
\subsection{Metrics}
We follow most previous works on referring and reasoning segmentation to evaluate model performance using two key metrics: \textit{gIoU} and \textit{cIoU}.

\textbf{Cumulative Intersection-over-Union (cIoU)} is computed by taking the cumulative intersection over the cumulative union of all predicted and ground-truth masks. Although this metric reflects a global measure of overlap across the entire dataset, it tends to be biased toward large-area objects and is sensitive to their presence, which might cause fluctuations in the metric's values.

\textbf{Generalized Intersection-over-Union (gIoU)} is defined as the average of per-image IoU scores across all samples. The metric treats each sample equally and offers a stable, fine-grained evaluation of the segmentation quality for both large and small objects.
\label{sec:metrics}
\subsection{Models and Training Setup}
To comprehensively evaluate the effectiveness of our proposed R2SM dataset, we benchmark five representative vision-language models for referring or reasoning expression segmentation: LISA~\cite{LaiTCLY0J24}, PSALM~\cite{ZhangMZB24}, ReLA~\cite{0072DJ23}, GLaMM~\cite{Rasheed0MS0CAX024}, and PixelLM~\cite{RenHW0FFJ24}. Each model is fine-tuned on different splits of the R2SM dataset for 10 epochs, where one epoch corresponds to a full pass through all queries in the respective split. The best-performing epoch is selected for evaluation. All experiments are conducted using a single 32GB V100 GPU. For detailed training configurations for each model, please refer to the appendix.

\subsection{Results}
We report segmentation results on three benchmark splits within the R2SM dataset: COCOA-cls, D2SA, and MUVA. Evaluations are conducted under two settings: \textit{All} (evaluated on both modal and amodal text queries) and \textit{Amodal} (tested only on amodal queries). The results include both baseline and R2SM-fine-tuned versions of representative vision-language models, with performance metrics measured using gIoU and cIoU as defined in Section~\ref{sec:metrics}.

As shown in Table~\ref{tab:exp}, fine-tuning on R2SM invariably improves performance across all models and dataset splits. Notably, GLaMM$^\dag$ achieves the highest scores on the COCOA-cls split, particularly in scenarios involving amodal occlusion. This strong performance can be attributed to GLaMM’s original design for the Grounded Conversation Generation (GCG) task, which requires detailed region understanding and precise pixel-level grounding. ReLA$^\dag$ stands out in the D2SA and MUVA splits, delivering strong results in amodal evaluation. Its region-based probability estimation aligns well with the textual context, enabling accurate segmentation in scenes with heavy occlusion.

We observe that on the COCOA-cls split, cIoU scores are consistently higher than gIoU, indicating that a small number of large-area objects dominate the cumulative overlap and obscure finer-grained performance differences. In contrast, gIoU surpasses cIoU on D2SA and MUVA, revealing that while the per-sample segmentation quality is reasonably high, the performance on large or high-resolution objects is less consistent. This discrepancy highlights the complementary nature of the two metrics.

In general, these results validate the effectiveness of the R2SM dataset in enhancing both modal and amodal segmentation capabilities. The gains across multiple models and splits underscore the role of R2SM as a strong benchmark for reasoning-based segmentation tasks.


\begin{table}[ht]
\caption{Performance comparison of various models on COCOA-cls, D2SA, and MUVA splits. Models marked with $^\dag$ are fine-tuned on the R2SM dataset, while those without are original baselines.}
\label{tab:exp}
\centering
\small
\setlength{\tabcolsep}{3.5pt}
\renewcommand{\arraystretch}{1.2}
\begin{tabular}{lccccccccccccc}
\toprule
\multirow{3}{*}{\textbf{Method}} 
& \multicolumn{4}{c}{\textbf{COCOA-cls Split}} 
& \multicolumn{4}{c}{\textbf{D2SA Split}} 
& \multicolumn{4}{c}{\textbf{MUVA Split}} \\
\cmidrule(lr){2-5} \cmidrule(lr){6-9} \cmidrule(lr){10-13}
& \multicolumn{2}{c}{All} & \multicolumn{2}{c}{Amodal} 
& \multicolumn{2}{c}{All} & \multicolumn{2}{c}{Amodal}
& \multicolumn{2}{c}{All} & \multicolumn{2}{c}{Amodal} \\
& gIoU & cIoU & gIoU & cIoU 
& gIoU & cIoU & gIoU & cIoU 
& gIoU & cIoU & gIoU & cIoU \\
\toprule
LISA-7B~\cite{LaiTCLY0J24} & 60.47 & 64.91 & 60.94 & 66.05 & 51.72 & 34.42 & 48.82 & 31.36 & 37.29 & 35.87 & 36.33 & 35.84 \\
\rowcolor{gray!15}
LISA-7B$^\dag$             & 65.08 & 71.81 & 65.61 & 72.88 & 75.78 & 71.21 & 77.03 & 73.47 & 54.52 & 53.77 & 55.21 & 54.55 \\
\midrule
PSALM~\cite{ZhangMZB24} & 59.55 & 58.11 & 59.02 & 58.50 & 44.40 & 21.96 & 43.45 & 21.97 & 30.18 & 19.68 & 31.54 & 23.07 \\
\rowcolor{gray!15}
PSALM$^\dag$            & 68.62 & 71.16 & 69.47 & 72.70 & 72.98 & 65.61 & 74.87 & 70.10 & 60.46 & 55.26 & 62.07 & 57.34 \\
\midrule
ReLA~\cite{0072DJ23}    & 45.31 & 51.01 & 44.58 & 50.70 & 30.65 & 22.31 & 29.61 & 20.72 & 24.88 & 23.22 & 23.38 & 21.89 \\
\rowcolor{gray!15}
ReLA$^\dag$             & 66.23 & 73.32 & 66.54 & 74.34 & \textbf{83.18} & \textbf{81.24} & \textbf{84.19} & \textbf{83.26} & \textbf{71.78} & \textbf{72.54} & \textbf{73.49} & \textbf{74.68} \\
\midrule
GLaMM~\cite{Rasheed0MS0CAX024} & 65.36 & 67.20 & 65.14 & 68.29 & 59.63 & 40.48 & 59.32 & 39.11 & 38.48 & 36.67 & 38.62 & 37.03 \\
\rowcolor{gray!15}
GLaMM$^\dag$                   & \textbf{70.73} & \textbf{75.56} & \textbf{71.88} & \textbf{77.14} & 78.79 & 74.35 & 80.00 & 76.51 & 59.30 & 58.34 & 60.29 & 59.29 \\
\midrule
PixelLM-7B~\cite{RenHW0FFJ24}     & 59.20 & 64.72 & 59.64 & 65.71 & 51.48 & 34.55 & 50.40 & 33.57 & 34.81 & 34.41 & 35.07 & 35.05 \\
\rowcolor{gray!15}
PixelLM-7B$^\dag$                 & 64.52 & 72.56 & 64.96 & 73.61 & 71.80 & 68.81 & 72.18 & 70.03 & 50.99 & 53.43 & 51.34 & 53.86 \\

\bottomrule
\end{tabular}
\end{table}

\begin{figure}[ht]
  \centering
  \includegraphics[width=\linewidth]{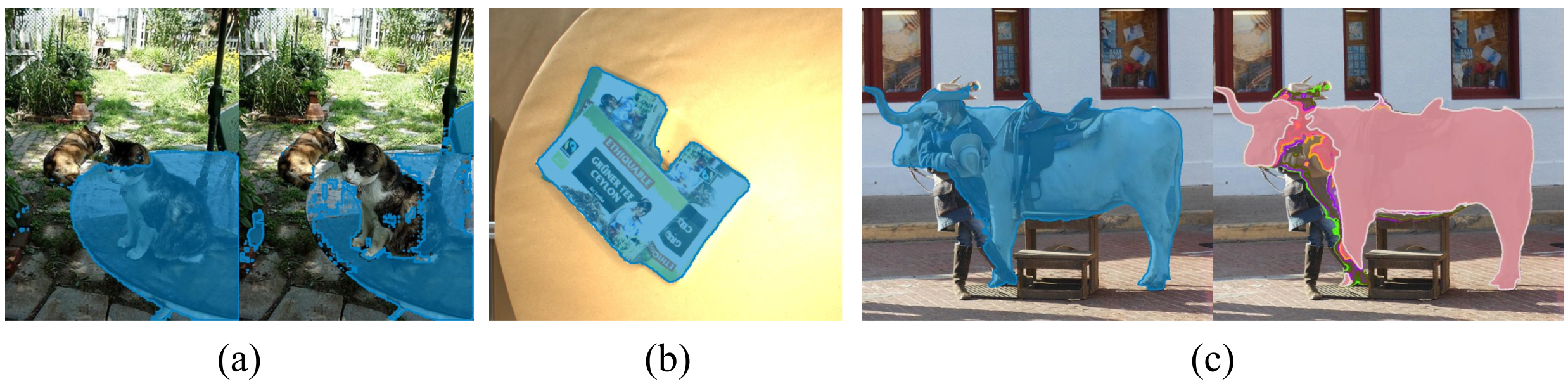}
  \caption{Examples of current limitations in R2SM, including (a) comparison between the results of amodal (left) and modal (right) query, (b) boundary ambiguity caused by overlapping objects, and (c) comparison of fused masks (left) and degraded instance-wise masks (right).}
  \label{fig:vis_limitation}
\end{figure}
\section{Challenges and Future Work}
Although fine-tuning on our R2SM dataset leads to notable improvements in quantitative results, qualitative analysis reveals that current models still struggle with key challenges. While they demonstrate some ability to differentiate between segmenting the visible part of an object and its complete, occluded shape based on the text prompt, visualizations expose persistent weaknesses in their predictions.

\subsection{Prompt Ambiguity}
Figure~\ref{fig:vis_limitation}(a) illustrates the predictions made by GLaMM. The left image is prompted to segment the complete table, including parts occluded by the cat, whereas the right image is prompted for the visible part only. The amodal prediction is moderately accurate, but the modal result shows signs of over-segmentation. This suggests that the model has difficulty distinguishing clearly between modal and amodal segmentation, occasionally blending reasoning strategies across prompts.
\subsection{Metric Misalignment}
Another challenge lies in the evaluation methodology for models that predict multiple instance masks per query, such as PSALM. Commonly used metrics like gIoU and cIoU are typically computed after fusing all predicted masks. In amodal scenarios where multiple queried objects overlap, as shown in Figure~\ref{fig:vis_limitation}(b), this fused evaluation obscures the performance of individual masks. A seemingly good overall segmentation might hide the fact that individual predictions are inaccurate or inconsistent.

To further investigate, we visualize the outputs of PSALM before and after mask fusion in Figure~\ref{fig:vis_limitation}(c). While the fused output (left) appears accurate, the original per-instance predictions (right) reveal noisy mask outputs, indicating a lack of precise instance-level understanding. These observations highlight the importance of using instance-level evaluation metrics, such as \textit{Average Precision (AP)} and \textit{Average Recall (AR)}, which are better suited to capture the performance of a model in fine-grained, occlusion-sensitive reasoning tasks.
\subsection{AP and AR Results}
We evaluate the performance of PSALM on the validation sets of all three R2SM splits using AP and AR. The evaluation follows the standard COCO evaluation protocol~\cite{LinMBHPRDZ14} with two key modifications. First, we treat each text query as an independent evaluation unit and match predictions to ground-truth masks by query ID, rather than by image ID. Second, since our task does not involve category prediction, we adopt a class-agnostic evaluation for AP and AR.

In Table~\ref{tab:ap_metrics}, \textit{All} refers to evaluations that include both modal and amodal queries, while the remaining rows correspond to amodal-only settings. Interestingly, although gIoU and cIoU reflect improved performance after fine-tuning when the evaluation is restricted to amodal queries, AP and AR exhibit a decline under the same condition.

This finding aligns with our qualitative analysis, where many individual predictions show low mask quality. This issue is often hidden when masks are fused for gIoU and cIoU evaluation. Such discrepancies highlight an important limitation in current evaluation methods and point to the need for more reliable, instance-aware evaluation protocols in future research.

\begin{table}[ht]
\caption{AP and AR results by PSALM on the R2SM dataset. AP$_s$, AP$_m$, and AP$_l$ refer to small ($< 32^2$ px), medium ($32^2$–$96^2$ px), and large ($\geq 96^2$ px) objects, respectively.}
\label{tab:ap_metrics}
\centering
\begin{tabular}{lcccccccccc}
\toprule
\textbf{Split} & \textbf{All} & \textbf{gIoU} & \textbf{cIoU} & \textbf{AP} & \textbf{AP$_{50}$} & \textbf{AP$_{75}$} & \textbf{AP$_s$} & \textbf{AP$_m$} & \textbf{AP$_l$} & \textbf{AR} \\
\midrule
\multirow{2}{*}{COCOA-cls} 
& \checkmark & 68.62 & 71.16 & 20.71 & 25.53 & 23.45 & 9.13 & 15.47 & 26.77 & 70.68 \\
& $\times$ & 69.47 & 72.70 & 18.66 & 23.55 & 20.68 & 8.72 & 12.86 & 24.50 & 68.06 \\
\midrule
\multirow{2}{*}{D2SA} 
& \checkmark & 72.98 & 65.61 & 34.62 & 37.93 & 36.31 & - & 8.67 & 34.69 & 84.87 \\

& $\times$ & 74.87 & 70.1 & 26.13 & 29.54 & 27.62 & - & - & 26.13 & 82.37 \\
\midrule
\multirow{2}{*}{MUVA} 
& \checkmark & 60.46 & 55.26 & 21.68 & 22.82 & 22.04 & 0.72 & 5.09 & 22.96 & 80.90 \\
& $\times$ & 62.07 & 57.34 & 20.50 & 22.18 & 20.96 & 2.22 & 3.40 & 20.84 & 76.77 \\
\bottomrule
\end{tabular}
\end{table}

\section{Conclusion}

In this paper, we introduce R2SM, a novel task and benchmark for evaluating vision-language models' ability to perform user-intent-aware segmentation through natural language. By combining modal and amodal segmentation within a unified framework, R2SM challenges models not only to understand visual content but also to reason about occlusion and select the appropriate mask type based solely on linguistic cues. We constructed natural language prompts for existing amodal datasets and associated them with ground-truth modal or amodal masks, enabling models to learn segmentation behavior conditioned on user intent. Through extensive experiments with five state-of-the-art models, we revealed significant gaps in their ability to address occlusion reasoning and intent-based segmentation. We hope R2SM will serve as a valuable resource for advancing multimodal understanding and developing more robust, intent-aware vision-language models.

\bibliographystyle{unsrtnat}
\bibliography{reference}

\clearpage

\appendix
\section{Training Details}

All models are fine-tuned separately on each R2SM split (COCOA-cls, D2SA, MUVA), and the best-performing checkpoint is selected based on gIoU on the validation set. Training is conducted on a single NVIDIA V100 GPU (32 GB memory). All models are initialized from the official pretrained weights released by their respective authors. For LISA, GLaMM, and PixelLM, we apply LoRA adapters with rank 8 and scaling factor 16. These three models update the segmentation and projection modules and LoRA-injected attention layers, while freezing the vision encoder and most of the language model.

ReLA, which uses a Swin Transformer-based image encoder and BERT for processing language expressions, is trained with a learning rate of $1 \times 10^{-5}$ and a batch size of 8, using a step learning rate scheduler. PSALM is fine-tuned with a learning rate of $6 \times 10^{-5}$ and a batch size of 8, using a cosine learning rate scheduler. Its backbone is frozen, and only the remaining components are updated; the best checkpoints are from epoch 10 (COCOA-cls) and epoch 9 (D2SA/MUVA).

LISA-7B is trained with a learning rate of $3 \times 10^{-4}$ and a batch size of 2. The best checkpoints are selected from epoch 9 (COCOA-cls/D2SA) and epoch 8 (MUVA). GLaMM is trained with a learning rate of $3 \times 10^{-4}$ and a batch size of 2, with the best checkpoints from epoch 6 across all splits. PixelLM is trained with a learning rate of $1 \times 10^{-5}$ and a batch size of 4, using the 7B variant of the language model. The best checkpoints are selected from epoch 10 for all splits.

\section{Qualitative Results}
Figure~\ref{fig:sup_cocoa}, Figure~\ref{fig:sup_d2sa}, and Figure~\ref{fig:sup_muva} present additional qualitative results from the COCOA, D2SA, and MUVA splits, respectively. We compare the segmentation outputs of five vision-language models: ReLA (yellow), LISA (green), PSALM (blue), GLaMM (purple), and PixelLM (red), each fine-tuned on R2SM. These examples demonstrate the models' ability to interpret natural language prompts and select between modal and amodal segmentation under varying levels of occlusion.

We observe that most models, after fine-tuning on the R2SM dataset, become capable of predicting the overall outline of an instance's amodal shape, even under partial occlusion. However, the generated amodal masks often exhibit coarse or imprecise edges, especially for the occluded regions, indicating difficulty in reconstructing fine details of the hidden parts. Additionally, we find that training with amodal annotations can negatively influence the quality of modal predictions. In some cases, the segmentation masks generated for visible instances become less precise, suggesting a trade-off between accurate occlusion reasoning and maintaining sharp modal boundaries.
\begin{figure}[ht]
  \centering
  \includegraphics[width=\linewidth]{figures_supp/cocoa.pdf}
  \caption{Qualitative results on the COCOA split. Each model’s output is shown with a background color: ReLA (yellow), LISA (green), PSALM (blue), GLaMM (purple), and PixelLM (red).}
  \label{fig:sup_cocoa}
\end{figure}
\begin{figure}[ht]
  \centering
  \includegraphics[width=\linewidth]{figures_supp/d2sa.pdf}
  \caption{Qualitative results on the D2SA split. Each model’s output is shown with a background color: ReLA (yellow), LISA (green), PSALM (blue), GLaMM (purple), and PixelLM (red).}
  \label{fig:sup_d2sa}
\end{figure}
\begin{figure}[ht]
  \centering
  \includegraphics[width=\linewidth]{figures_supp/muva.pdf}
  \caption{Qualitative results on the MUVA split. Each model’s output is shown with a background color: ReLA (yellow), LISA (green), PSALM (blue), GLaMM (purple), and PixelLM (red).}
  \label{fig:sup_muva}
\end{figure}
\section{R2SM Dataset Generation Prompts}
\begin{figure}[ht]
  \centering
  \includegraphics[width=\linewidth]{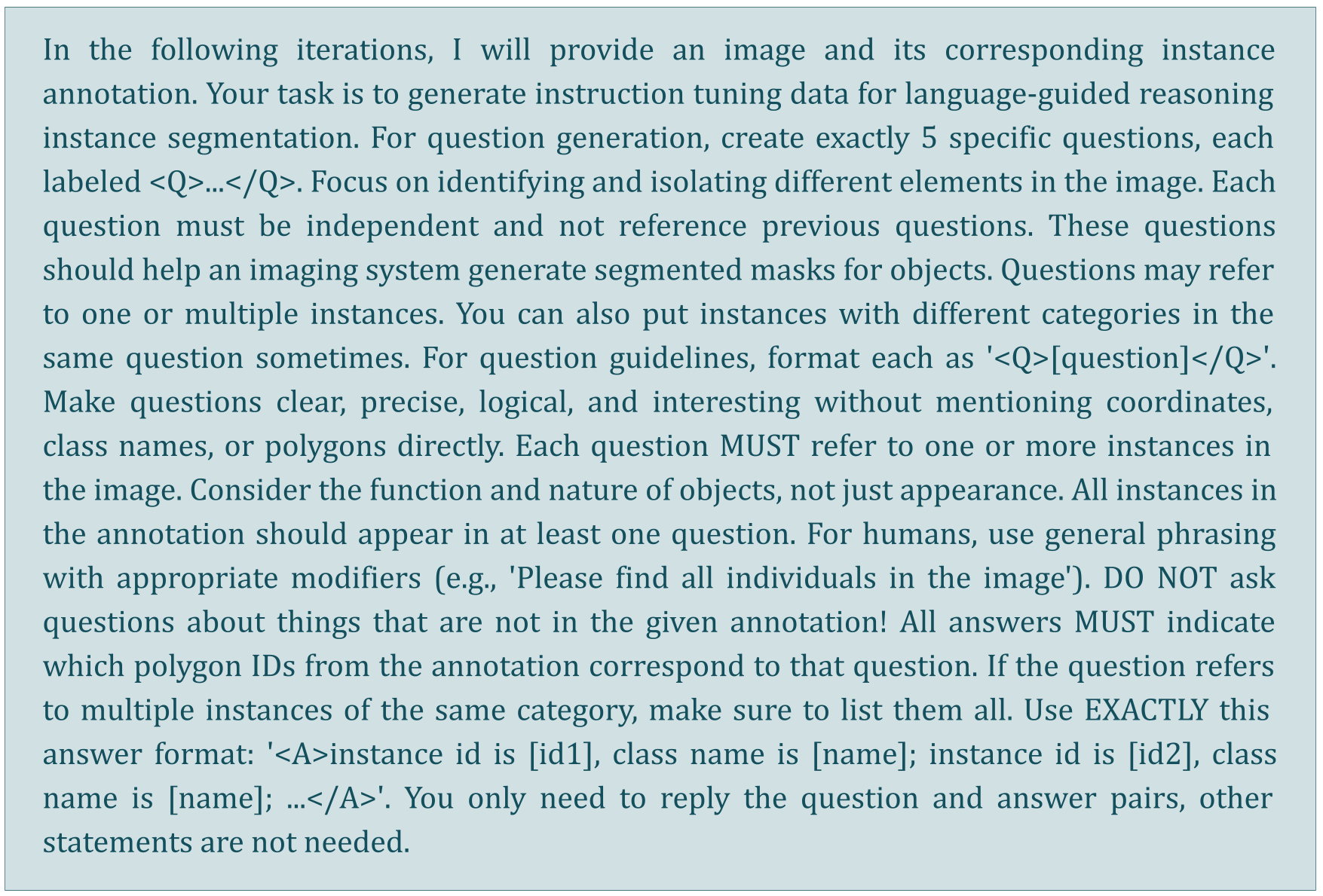}
  \caption{System prompt template provided to the Claude API for generating modal text queries in the R2SM dataset.}
  \label{fig:sup_modal}
\end{figure}

\begin{figure}[ht]
  \centering
  \includegraphics[width=\linewidth]{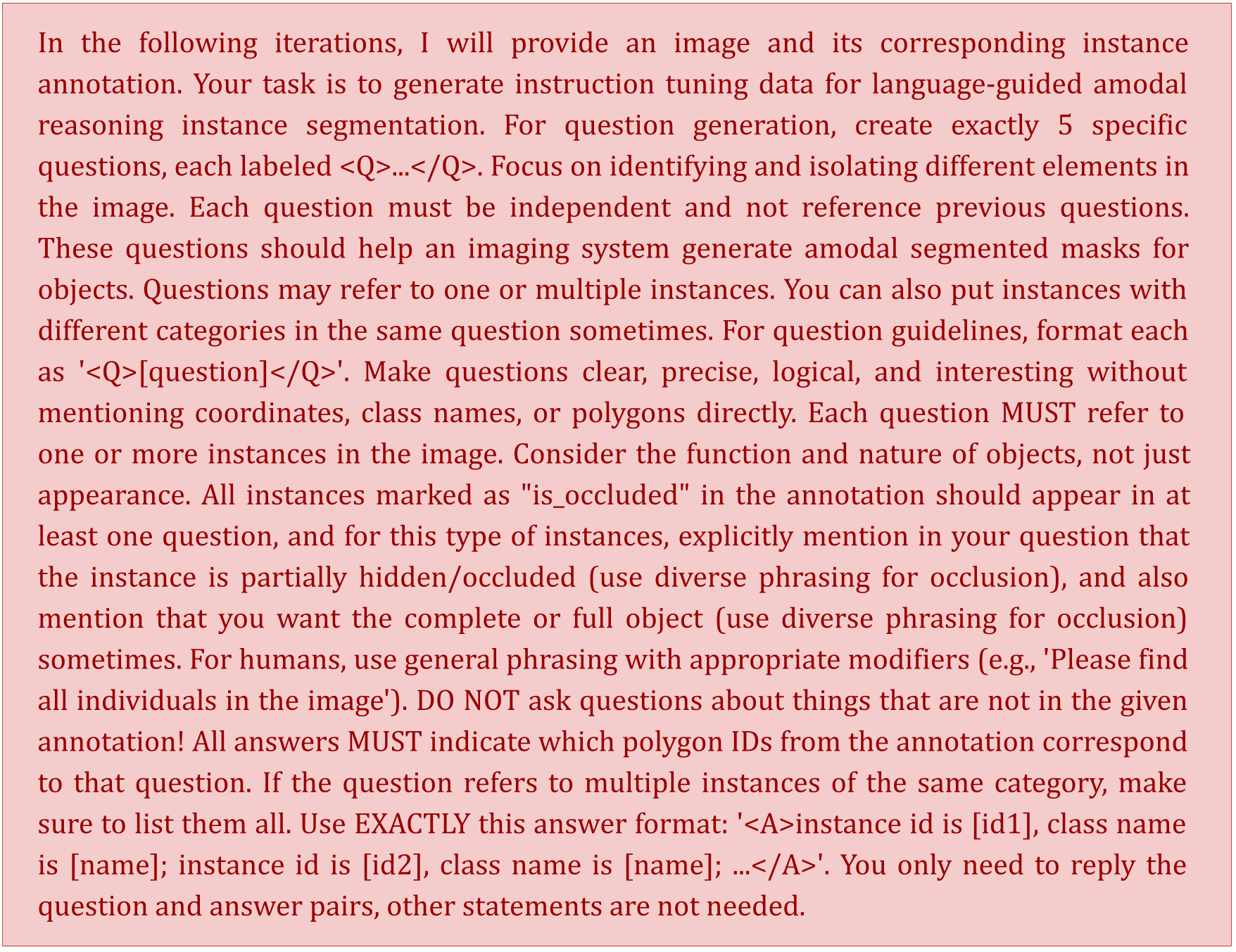}
  \caption{System prompt template provided to the Claude API for generating amodal text queries in the R2SM dataset. Instances marked with \textit{is\_occluded} are explicitly highlighted, and Claude is instructed to generate descriptions that account for occlusion and infer the complete object shape beyond visible regions.}
  \label{fig:sup_amodal}
\end{figure}

\begin{figure}[ht]
  \centering
  \includegraphics[width=\linewidth]{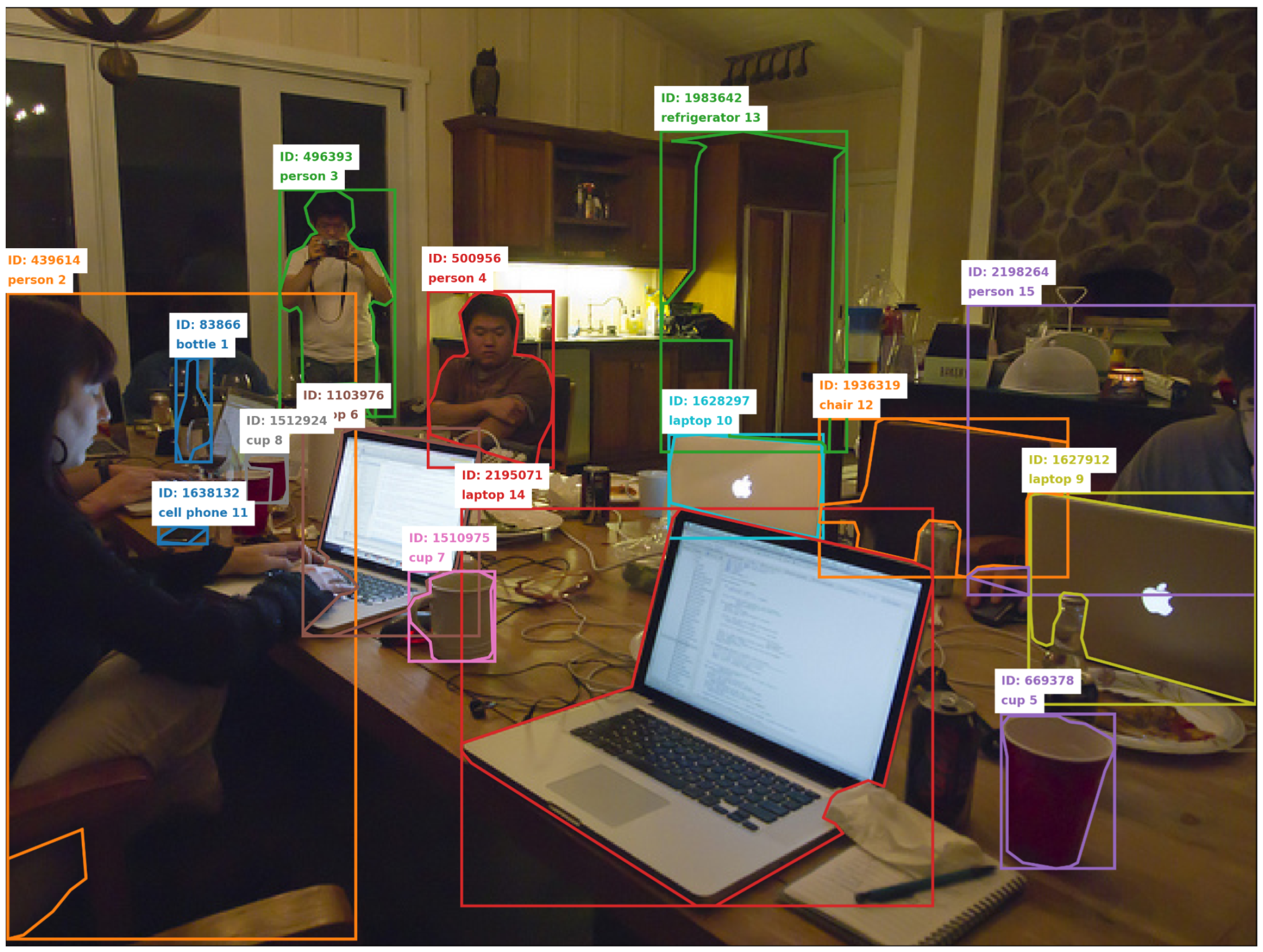}
  \caption{Example image provided to the Claude API during the text query generation process. To help Claude understand the expected output format and context, we supply reference images along with structured annotations and sample question-answer pairs.}
  \label{fig:sup_samp_img}
\end{figure}

\begin{figure}[ht]
  \centering
  \includegraphics[width=\linewidth]{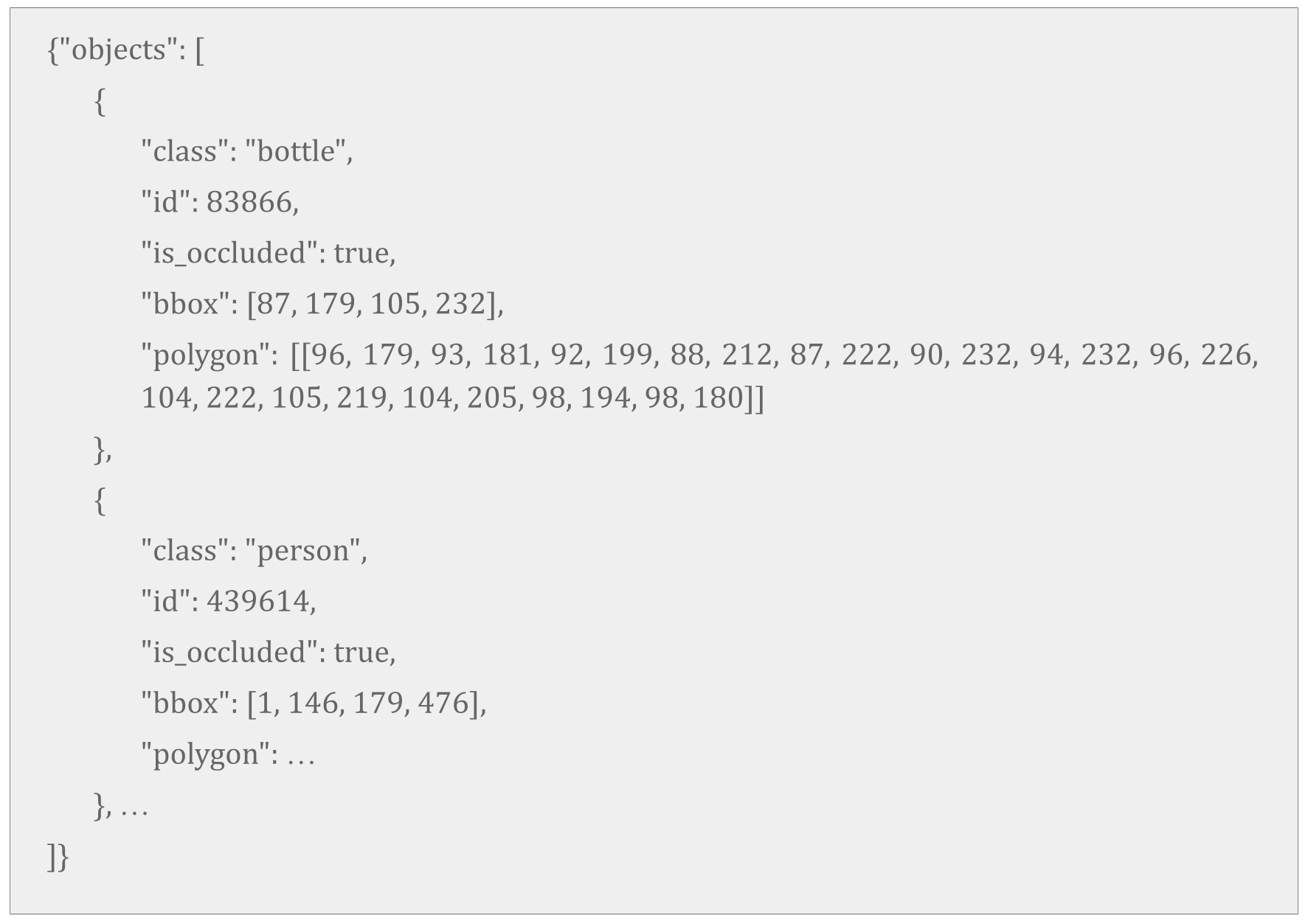}
  \caption{Instance-level annotation example, including category labels, bounding boxes, segmentation masks, and occlusion flags for all objects in the image.}
  \label{fig:sup_samp_ann}
\end{figure}

\begin{figure}[ht]
  \centering
  \includegraphics[width=\linewidth]{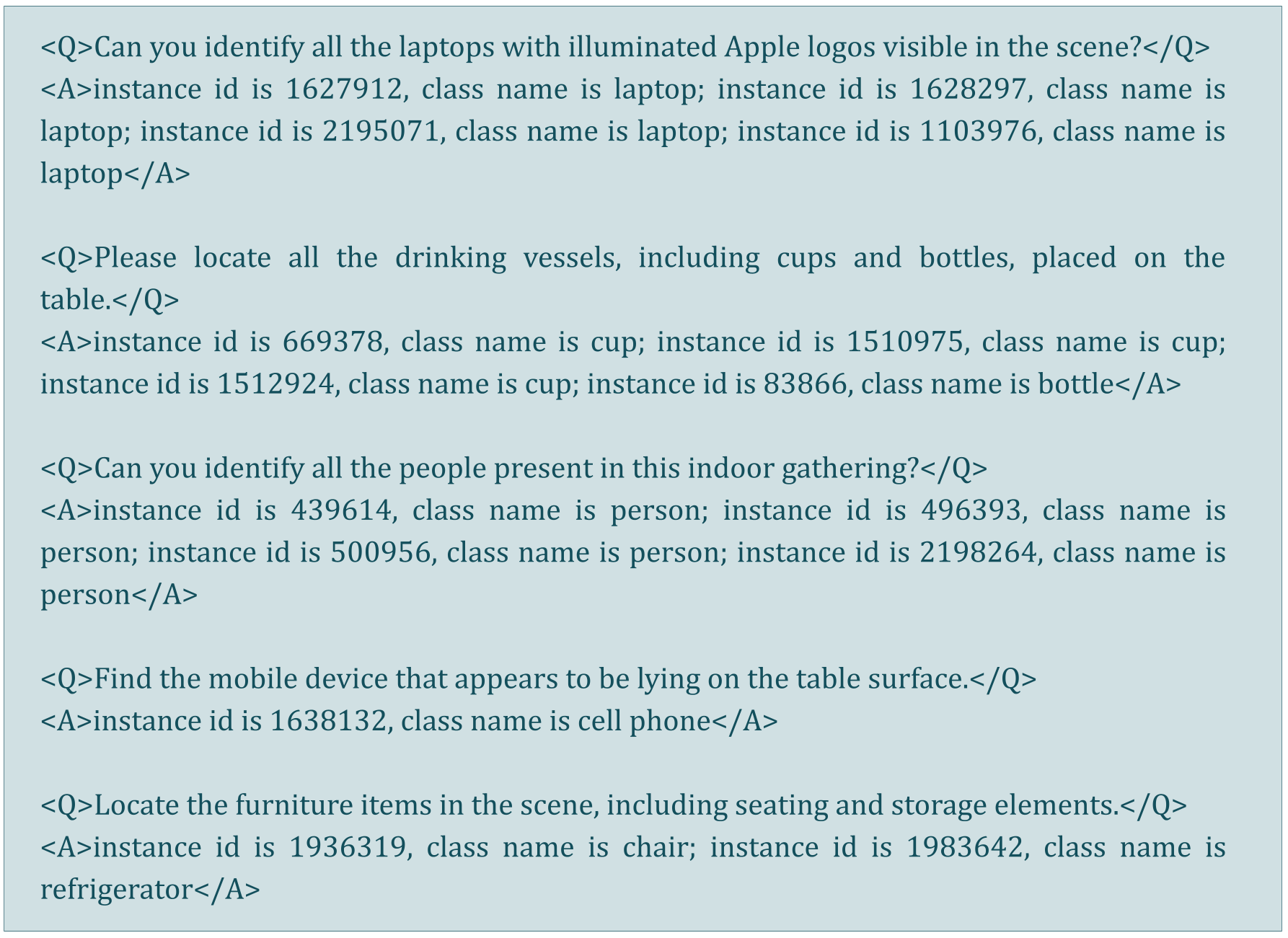}
  \caption{In-context sample of modal question-answer pairs included in the prompt to guide Claude’s generation.}
  \label{fig:sup_samp_mod}
\end{figure}

\begin{figure}[ht]
  \centering
  \includegraphics[width=\linewidth]{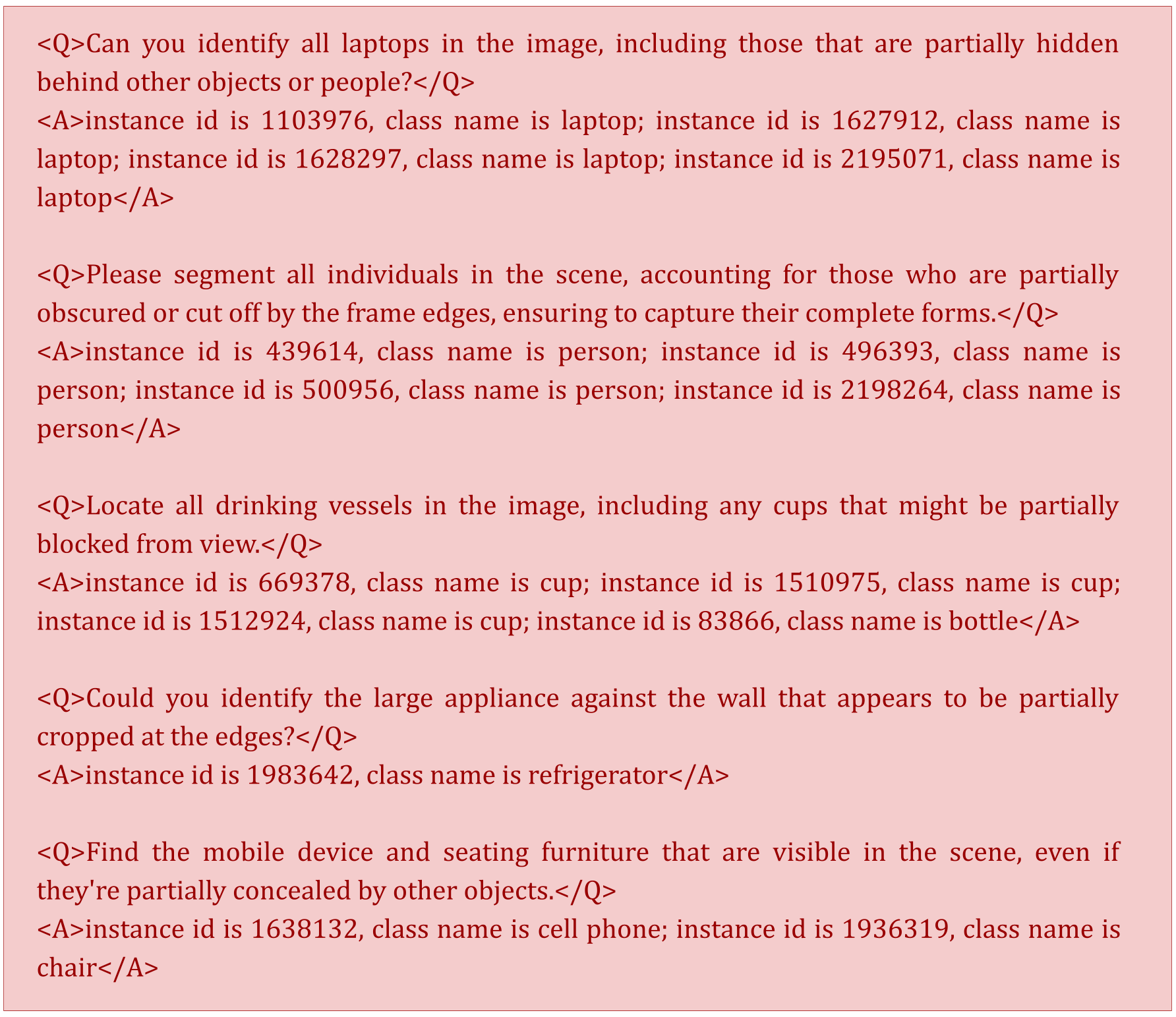}
  \caption{In-context sample of amodal question-answer pairs included in the prompt to guide Claude’s generation. The amodal queries emphasize object occlusion and encourage complete shape inference.}
  \label{fig:sup_samp_amod}
\end{figure}

Figure~\ref{fig:sup_modal} and Figure~\ref{fig:sup_amodal} show the system prompts used to generate modal and amodal queries using the Claude API. To condition the language model effectively, we provide instance-level annotations including category names, bounding boxes, segmentation masks, and occlusion flags, as shown in Figure~\ref{fig:sup_samp_img} and Figure~\ref{fig:sup_samp_ann}. Figures~\ref{fig:sup_samp_mod} and Figure~\ref{fig:sup_samp_amod} illustrate the in-context examples embedded in the prompt to constrain Claude’s response style and structure for generating modal and amodal queries.




\clearpage

\end{document}